\documentclass[runningheads]{llncs}
\usepackage[T1]{fontenc}
\usepackage{graphicx,verbatim}
\usepackage{amssymb}

\usepackage{booktabs}
\usepackage{multirow}
\usepackage{makecell}
\usepackage{amsmath}
\usepackage{booktabs}
\usepackage[section]{placeins}

\usepackage{pgfplots}
\usepackage{tikz}
\usetikzlibrary{pgfplots.groupplots,patterns}
\pgfplotsset{compat=1.18}

\begin{document}
\title{Program-space Diffusion for Morphology-to-Transcriptomics Prediction }
\titlerunning{Program-space Diffusion for Morphology-to-Transcriptomics Prediction}
%

\author{
Swann Ruyter\inst{1}$^{*}$
\and
Reuben Dorent\inst{1}$^{\dagger}$
\and
Daniel Racoceanu\inst{1}$^{\dagger}$
}

\authorrunning{S. Ruyter et al.}

\institute{
Sorbonne Universit\'e, CNRS, Inserm, AP-HP, Inria, Paris Brain Institute -- ICM, Paris, France\\
\email{swann.ruyter@icm-institute.org},
\email{reuben.dorent@inria.fr},
\email{daniel.racoceanu@sorbonne-universite.fr}\\
$^{*}$Corresponding author. \quad
$^{\dagger}$These authors contributed equally.
}

  
\maketitle              
\begin{abstract}
Spatial transcriptomics (ST) enables genome-wide gene expression profiling while preserving tissue architecture, but its cost and limited scalability remain major bottlenecks. This has motivated models that predict spatial expression directly from routine histology. Despite promising results, most existing approaches operate at the gene level without leveraging established transcriptomic modeling practices and rely on heterogeneous gene selection strategies, which complicates fair comparison across methods.

We propose to reformulate morphology-to-transcriptomics prediction as conditional generation in transcriptional program space, thereby exploiting coordinated transcriptional variation instead of predicting genes independently. Using consensus non-negative matrix factorization (cNMF), we extract a low-dimensional set of transcriptional programs capturing coordinated expression variation in the training data, and train a conditional diffusion model to generate program activations from histology. This formulation exploits coordinated transcriptional variation and substantially lowers the dimensionality of the conditional generative task.


\end{abstract}

\section{Introduction}

Spatial transcriptomics (ST) enables \emph{in situ} gene expression profiling while preserving tissue architecture, thereby establishing a direct link between routine histology (H\&E whole-slide imaging, WSI) and molecular states. This joint view allows morphological patterns observed in WSI to be interpreted in the context of spatially resolved transcriptional variation, providing a more comprehensive characterization of tissue organization than either modality alone \cite{rao_exploring_2021}.
However, in practice, ST remains costly and difficult to acquire, which motivates the development of methods that predict spatial gene expression directly from widely available H\&E slides. This task is inherently challenging due to the noisy and technology-dependent nature of ST measurements (sequencing depth, technical noise, dropouts, and sectioning or image–spot alignment artifacts), as well as the fundamental non-uniqueness of the morphology–expression relationship \cite{du_advances_2023,you_systematic_2024}.


Early \emph{morpho-transcriptomic} models largely framed the problem as supervised regression by pairing H\&E patches centered at spot locations with spatial gene expression vectors \cite{he_integrating_2020}. Subsequent work changed network architectures, moving from CNN-based encoders to multi-scale Transformers and spatial reasoning modules such as graph neural networks to better exploit tissue context \cite{pang_leveraging_2021,zeng_spatial_2022,monjo_efficient_2022,yang_spatial_2023}.
However, these methods do not account for non-uniqueness of the morphology–expression relationship, where multiple transcriptional states can be compatible with similar histology. To address this ambiguity, STEM~\cite{zhu_diffusion_2025} introduced a probabilistic formulation based on conditional diffusion and reported strong performance.
Yet, two key limitations remain. First, STEM relies on log-transformed ST counts, which compress the dynamic range of expression values but do not fully account for sequencing-depth effects or the intrinsic mean–variance dependency characteristic of count data \cite{ruyter_normalization_nodate}. Consequently, residual technical variability may remain entangled with biological signal in gene space. Second, STEM’s computational cost scales poorly with the number of genes being predicted. This limitation is particularly important, as transcriptional variation is highly correlated across genes, and constraining inference to a small, predefined gene set may bias both model learning and downstream biological interpretation.

In this work, we address these two limitations of diffusion-based morpho-transcriptomic inference. First, we adopt an analytically grounded normalization strategy for spatial transcriptomics data based on Pearson residuals derived from a parsimonious count model~\cite{lause_analytic_2021}. Unlike standard log-based preprocessing, this approach explicitly accounts for sequencing depth and the mean--variance relationship of UMI counts, yielding approximately variance-stabilized representations. Second, we reformulate morphology-to-transcriptomics prediction in a structured program space learned by consensus non-negative matrix factorization (cNMF), simultaneously improving computational scalability while constraining predictions to coordinated transcriptional programs. 
cNMF has previously been used to identify reproducible transcriptional programs in single-cell analysis \cite{kotliar_identifying_2019}.
Specifically, this data-driven, structured representation constrains predictions to linear combinations of transcriptional programs rather than independent genes, while significantly reducing computational cost. Third, controlled experiments on a public HER2+ cohort with patient-held-out splits demonstrate (i) the substantial impact of normalization on predictive performance, (ii) improved per-slide predictive quality with larger gene panels, with moderate gains in fixed-K program space, and (iii) the scalability of our proposed program-based approach.

\section{Method}

\subsection{Shifted Pearson residual normalization}\label{sec:pearson_normalization}
To normalize spatial transcriptomics expression, we employ a variance-stabilized transformation based on Pearson residuals derived from a parsimonious count model~\cite{lause_analytic_2021}. This normalization explicitly accounts for spot-specific sequencing depth and the mean-variance relationship intrinsic to unique molecular identifiers (UMI) count data, leading to expression values that are less affected by technical variability and more amenable to downstream modeling.

Let $X \in \mathbb{R}_{\ge 0}^{N \times G}$ be a raw count matrix, where  $N$ and $G$ are respectively the number of spots and genes. We compute Pearson residuals under a negative binomial (NB) model with fixed dispersion $\theta$.
First, we define the library size $n_i =~\sum_{j} X_{ij}$, the gene proportion
$p_j = \frac{\sum_{i} X_{ij}}{\sum_{i,j} X_{ij}}$, and the expected mean $\mu_{ij} =~n_i p_j$.
Then, Pearson residuals are computed as
\begin{equation}
R_{ij} \;=\; \frac{X_{ij} - \mu_{ij}}{\sqrt{\mu_{ij} + \mu_{ij}^2 / \theta}} \, ,
\end{equation}
where $\theta$ denotes a fixed dispersion parameter and a numerical floor of $10^{-8}$ is applied to the denominator for stability. This formulation leads to approximately variance-stabilized residuals across genes and expression levels.

To further limit the influence of extreme values, residuals are clipped element-wise to the interval $[-c, c]$. Since cNMF and non-negative least squares require non-negative inputs, we apply a global shift to the residuals:
\begin{equation}
\tilde{\mathbf{X}} = R - \min(R) + \varepsilon,
\label{eq:definition_x_norm}
\end{equation}
where $\varepsilon > 0$. In practice, we set $\theta = 100$, $c = 10$, and $\varepsilon = 10^{-6}$.

Because gene-wise proportions $p_j$ are estimated on the selected gene panel, this normalization is panel-dependent; the panel is therefore fixed prior to normalization to ensure consistency across downstream modeling steps.

\subsection{Consensus Non-negative Matrix Factorization (cNMF)}
\label{sec:cnmf}

To obtain a compact
representation of transcriptional variation, we construct a set of transcriptional programs via consensus non-negative matrix factorization (cNMF) within each training fold. Each spot's expression profile is then represented as a non-negative linear combination of these programs.

\paragraph{Multiple NMF runs.}
cNMF is applied to the normalized expression matrix $\tilde{\mathbf{X}}$ defined in \eqref{eq:definition_x_norm}. 
To improve stability and mitigate sensitivity to initialization, we follow the consensus strategy of \cite{kotliar_identifying_2019} and perform $R=20$ independent NMF runs. The scikit-learn implementation was used with NNDSVDa initialization (\texttt{init=nndsvda}), coordinate descent (\texttt{solver=cd}) and Frobenius reconstruction objective (\texttt{beta\_loss=frobenius}). Each run $r$ produces a factorization
\begin{equation}
\tilde{\mathbf{X}} \approx \mathbf{H}^{(r)} \mathbf{W}^{(r)},
\end{equation}
where $\mathbf{H}^{(r)} \in \mathbb{R}_{\ge 0}^{N \times K}$ contains spot-level activations for the $K$ programs and $\mathbf{W}^{(r)} \in \mathbb{R}_{\ge 0}^{K \times G}$ contains gene loadings.  The number of programs $K$ controls the trade-off between reconstruction fidelity and dimensionality reduction. 
To examine sensitivity to $K$, we performed cNMF across $K \in \{10, 20, 40, 60, 100\}$ in a patient-held-out setting. Increasing $K$ improved oracle reconstruction quality but produced denser per-spot activations. We therefore use $K=20$ as a compact operating point balancing reconstruction fidelity, activation sparsity, and target dimensionality; it is not interpreted as a biologically optimal or universally transferable rank.

\paragraph{Consensus programs.}
To obtain stable programs across runs, we aggregate the loading matrices $\{\mathbf{W}^{(r)}\}_{r=1}^R$ by stacking them into a matrix of shape $(RK) \times G$ and clustering rows using $K$-means with $K$ clusters (default \texttt{n\_init=20}). Prior to clustering, each row of $\mathbf{W}^{(r)}$ is $\ell_1$-normalized to ensure comparability across runs. Programs assigned to the same cluster are averaged to form the consensus loading matrix $\mathbf{W}^* \in \mathbb{R}_{\ge 0}^{K \times G}$, and each consensus program is subsequently $\ell_1$-normalized. The resulting basis $W^*$ defines a consensus set of transcriptional programs representing coordinated expression variation.

\paragraph{Projection via NNLS.}
Given the fixed consensus basis $\mathbf{W}^*$, spot-level activations are estimated via non-negative least squares:
\begin{equation}
\mathbf{H} \;=\; \arg\min_{\mathbf{H} \ge 0} 
\| \tilde{\mathbf{X}} - \mathbf{H} \mathbf{W}^* \|_F^2.
\label{eq:decomposition_optimization}
\end{equation}

\noindent Consequently, the diffusion model predicts only the low-dimensional activation coefficients $\mathbf{H}$, while gene expression is approximated within the subspace spanned by the fixed basis $\mathbf{W}^*$.

\subsection{Conditional Program-Space Diffusion}
\label{sec:diffusion_program}

Given a histology image patch $\mathbf{I}_i$ at spatial location $i$, the objective is to predict the corresponding gene expression profile $\mathbf{\tilde{X}}_i$. Using the cNMF decomposition in~\eqref{eq:decomposition_optimization}, the gene expression profile $\mathbf{\tilde{X}}_i$ can be approximated as
\begin{equation}
    \mathbf{\tilde{X}}_i \approx \mathbf{h}_i \mathbf{W}^*,
\end{equation}
with $\mathbf{h}^{(0)}_i = \mathbf{H}_i \in \mathbb{R}^{1 \times K}$ denoting the spot-level program activations and $\mathbf{W}^*$ the fixed consensus basis. Rather than predicting individual genes directly, we therefore aim to predict the program decomposition $\mathbf{h}_i^{(0)}$, from which gene expression can be approximated through the fixed consensus basis.

We adopt a conditional diffusion model in program space to model the conditional distribution of program activations given histology. Specifically, we approximate the conditional distribution $
p_{\theta}(\mathbf{h}^{(0)}_i \mid \mathbf{I}_i)$, which generates spot-level activations $\hat{\mathbf{h}}_i$ given a histology patch $\mathbf{I}_i$.

\paragraph{Target space.}
We follow the discrete-time DDPM formulation with an $\epsilon$-prediction objective \cite{ho_denoising_2020}. The key modification relative to gene-space approaches is that diffusion operates in $\mathbb{R}^K$ rather than $\mathbb{R}^G$, with $K \ll G$. This reduces the dimensionality of the generative task and constrains predictions to coordinated transcriptional programs through the basis $\mathbf{W}^*$.  The consensus basis $\mathbf{W}^*$ is treated as fixed within each training fold and is not updated during diffusion training.

\paragraph{Forward process.}

During training, the forward diffusion process gradually perturbs program activations $\mathbf{h}_i$ with Gaussian noise over $T$ timesteps:
\begin{equation}
q(\mathbf{h}^{(t)}_i \mid \mathbf{h}^{(0)}_i)
=
\mathcal{N}\!\left(
\sqrt{\bar{\alpha}_t}\,\mathbf{h}^{(0)}_i,\;
(1-\bar{\alpha}_t)\mathbf{I}
\right),
\end{equation}
where $\bar{\alpha}_t = \prod_{s=1}^{t} (1-\beta_s)$ and $\{\beta_t\}_{t=1}^{T}$ is a predefined noise schedule. Equivalently,
\begin{equation}
\mathbf{h}^{(t)}_i
=
\sqrt{\bar{\alpha}_t}\,\mathbf{h}^{(0)}_i
+
\sqrt{1-\bar{\alpha}_t}\,\boldsymbol{\varepsilon},
\qquad
\boldsymbol{\varepsilon} \sim \mathcal{N}(\mathbf{0},\mathbf{I}).
\end{equation}
As $t$ increases, $\mathbf{h}^{(t)}_i$ progressively loses information about $\mathbf{h}^{(0)}_i$ and approaches isotropic Gaussian noise.

\paragraph{Reverse process and training.}

A neural network $\boldsymbol{\varepsilon}_{\theta}$ is trained to predict the injected noise at each timestep, conditioned on the noisy program activations, the corresponding histology patch $\mathbf{I}_i$, and the timestep $t$:
\begin{equation}
\mathcal{L}_{\mathrm{diff}}
=
\mathbb{E}_{\mathbf{h}^{(0)}_i,\boldsymbol{\varepsilon},t}
\left[
\left\|
\boldsymbol{\varepsilon}
-
\boldsymbol{\varepsilon}_{\theta}(\mathbf{h}^{(t)}_i, \mathbf{I}_i, t)
\right\|_2^2
\right].
\end{equation}
Minimizing $\mathcal{L}_{\mathrm{diff}}$ enables the model to learn the reverse denoising dynamics.

\paragraph{Reconstruction.}

Here, reconstruction denotes projection back to gene space through the fixed training-fold cNMF basis and does not imply recovery of expression components outside this subspace. At inference time, program activations $\hat{\mathbf{h}}^{(0)}_i$ are obtained by iteratively applying the learned reverse updates, starting from an isotropic Gaussian noise initialization $\hat{\mathbf{h}}^{(T)}_i \sim \mathcal{N}(\mathbf{0}, \mathbf{I})$. Gene expression is then reconstructed through the fixed consensus basis:
\begin{equation}
\hat{\mathbf{X}}_i = \hat{\mathbf{h}}^{(0)}_i \mathbf{W}^*.
\end{equation}

\subsection{Implementation details}

We adopt the standard DDPM diffusion parameterization as implemented in STEM \cite{zhu_diffusion_2025}. For all experiments, we use the default STEM diffusion configuration (noise schedule and timestep discretization). Sampling is performed conditionally on the histology embedding on GPU (CUDA) with $1000$ reverse-diffusion steps, generating $20$ samples per condition in batches of $200$. Training was run on a single 80\,GB NVIDIA H100 GPU.

\section{Experiments}

\subsubsection{Dataset}
We evaluate our framework on the publicly available HER2-positive breast cancer spatial transcriptomics cohort \cite{andersson_spatial_2021}, generated using first-generation Spatial Transcriptomics arrays. The dataset contains 36 tissue sections from eight patients (35 sections retained after quality control), with approximately 300 to 700 spots per slide.
Each section is paired with H\&E WSI, spot coordinates, and spot-level gene expression.
For every spot, we extract a fixed-size image patch centered at the provided coordinates and pair it with its corresponding expression vector harmonized by \cite{jaume_hest-1k_2024} in HEST-1K benchmark.
All experiments follow a patient-held-out cross-validation scheme, ensuring that all slides from the held-out patient are excluded from training. All decompositions and projections are performed within each patient-held-out fold: the cNMF basis is learned exclusively from training-patient slides, and slides from the held-out patient are projected using NNLS to prevent information leakage. Consequently, this evaluation assesses generalization to held-out patients within the same cohort, but not transfer to independent cohorts or newer ST platforms.

\subsubsection{Gene selection} 
We follow the STEM gene selection procedure: Highly Variable Genes (HVGs) are first identified independently within each slide  and aggregated via union across slides. The resulting pool is then globally prioritized by combining mean expression and dispersion computed on the concatenated dataset, retaining genes that exhibit both high abundance and high variability. This hybrid strategy integrates local variability with global statistical constraints. From this procedure, we derive panels of $G\in\{300,800,2000\}$ genes.

\subsubsection{Evaluation protocols (normalization and splits)}
Normalization affects gene-level metrics and can confound method comparisons. We therefore report results under two complementary evaluation protocols.
\textit{Protocol A: log normalization.} STEM is kept under its original log-normalization, i.e.\ an element-wise $\log_{2}(X+1)$ transform without library-size normalization, to preserve the conditions under which it was designed and validated.
\textit{Protocol B: pearson normalization.} We additionally adopt the Pearson-residual normalization as described in~\ref{sec:pearson_normalization}.
In this setting, our cNMF program extraction and program-space diffusion operate on Pearson-residual normalization.
To isolate the impact of normalization, we keep gene panels, splits, architectures, checkpoints, and evaluation metrics fixed, and vary only the normalization.
As an explicit control, we also report STEM performance under Pearson-residual normalization (gene-space, $G=2000$) using identical splits and evaluation settings, so that any change can be attributed to normalization rather than modeling choices.

\begin{table}[t]
\centering
\caption{\textbf{Top-$M$ PCC (\%) for STEM-2000.}
Genes are ranked by PCC and Top-$M$ is the mean PCC over the top $M$ genes; reported values are averaged across slides.
Pearson Norm. indicates Pearson-residual space ($\checkmark$) vs.\ STEM's normalization ($\times$). Bold indicates the best predictive result under matched Pearson-residual normalization; the cNMF Oracle is reported only as a reconstruction diagnostic.}
\label{tab:stem_topm_pcc_N2000}
\begin{tabular}{@{}lcccccccccc@{}}
\toprule
\textbf{Setups} &
\makecell{\textbf{Pearson}\\\textbf{Norm.}} &
\multicolumn{9}{c}{\textbf{Top-$M$ PCCs}\,$\uparrow$ (\%)} \\
\cmidrule(lr){3-11}
& & 10 & 20 & 50 & 100 & 300 & 600 & 800 & 1500 & 2000 \\
\midrule

STEM \cite{zhu_diffusion_2025} & $\times$ & 53.2 & 50.4 & 46.2 & 42.7 & 36.7 & 32.5 & 30.5 & 25.1 & 20.9 \\
\midrule
STEM \cite{zhu_diffusion_2025} & $\checkmark$ & 46.5 & 41.8 & 34.8 & 29.3 & 20.8 & 15.5 & 13.3 & 8.1 & 4.8 \\
Ours & $\checkmark$ & \textbf{47.7} & \textbf{43.6} & \textbf{37.1} & \textbf{31.6} & \textbf{22.6} & \textbf{17.0} & \textbf{14.6} & \textbf{9.0} & \textbf{5.4} \\
\midrule
\textit{cNMF Oracle NNLS}  & $\checkmark$  & \textit{75.1} & \textit{70.2} & \textit{62.2} & \textit{55.3} & \textit{43.1} & \textit{35.0} & \textit{31.5} & \textit{23.4} & \textit{18.4} \\

\bottomrule
\end{tabular}
\end{table}

\begin{figure}[t!]
    \centering
    \includegraphics[width=\linewidth]{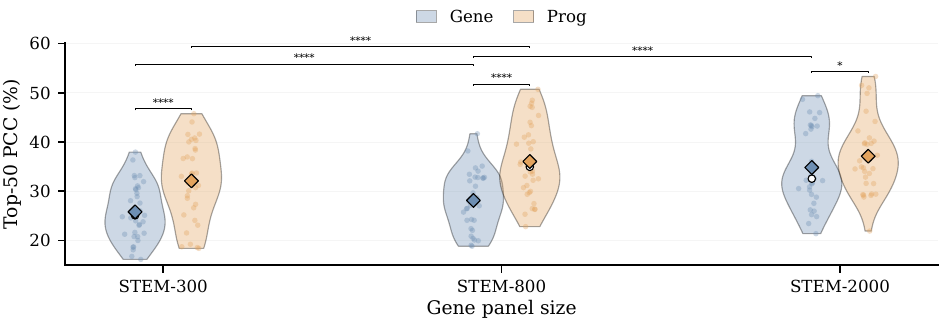}
    \caption{\textbf{Top-50 PCC per slide.} Distribution of per-slide Top-50 PCC scores under patient-held-out for gene-space (Gene) and program-space (Prog) models across gene panels  $G$ = {300, 800, 2000}. Significance assessed with paired Wilcoxon signed-rank tests.}
    \label{fig:violin_top50_paper}

\definecolor{GeneBlue}{RGB}{0,114,178}
\definecolor{ProgRed}{RGB}{213,94,0}

\FloatBarrier
\centering
\begin{tikzpicture}

\pgfplotsset{
  geneBar/.style={ybar, bar width=8pt, draw=black, fill=black!20},
  progBar/.style={ybar, bar width=8pt, draw=black, fill=white,
                  pattern=north east lines, pattern color=black},
  genePt/.style={only marks, mark=*, mark size=2.3pt,
                 draw=GeneBlue, fill=GeneBlue, mark options={solid}},
  progPt/.style={only marks, mark=square*, mark size=2.3pt,
                 draw=ProgRed, fill=ProgRed, mark options={solid}}
}

\begin{groupplot}[
  group style={group size=2 by 1, horizontal sep=1.7cm},
  width=0.52\textwidth,
  height=0.36\textwidth,
  xtick={1,2,3},
  xticklabels={STEM-300,STEM-800,STEM-2000},
  xmin=0.5, xmax=3.5,
  tick label style={font=\scriptsize},
  label style={font=\scriptsize},
  title style={font=\scriptsize},
  ymajorgrids,
  grid style={line width=.1pt, draw=black!12},
  clip=false
]

\nextgroupplot[
  title={Training (5k steps)},
  ylabel={minutes},
  ymin=0, ymax=105
]
\addplot+[geneBar] coordinates {(0.85,25) (1.85,91) (2.85,83)};
\addplot+[progBar] coordinates {(1.15,8)  (2.15,8)  (3.15,8)};
\addplot+[genePt]  coordinates {(0.85,25) (1.85,91) (2.85,83)};
\addplot+[progPt]  coordinates {(1.15,8)  (2.15,8)  (3.15,8)};

\node[font=\bfseries\scriptsize, anchor=east] at (rel axis cs:-0.10,1.20) {A};
\node[font=\scriptsize, anchor=south] at (axis cs:2.85,83) {bs=96};

\nextgroupplot[
  title={Sampling time (s, 1k steps)},
  ylabel={seconds},
  ymin=-100, ymax=1100,
  ytick={0,500,1000}
]

\addplot+[genePt] coordinates {(0.85,36.63) (1.85,135.14) (2.85,1000.00)};
\addplot+[progPt] coordinates {(1.15,20) (2.15,20) (3.15,20)};

\node[font=\bfseries\scriptsize, anchor=east] at (rel axis cs:-0.10,1.20) {B};

\end{groupplot}

\node[anchor=north, yshift=-6mm] at
  ($(group c1r1.south west)!0.5!(group c2r1.south east)$) {%
  \scriptsize
  \tikz[baseline=-0.6ex]\filldraw[draw=GeneBlue, fill=GeneBlue] (0,0) circle (1.6pt);
  \, gene-space
  \hspace{1.4em}
  \tikz[baseline=-0.6ex]\filldraw[draw=ProgRed, fill=ProgRed] (0,0) rectangle (3.2pt,3.2pt);
  \, program-space ($K = 20$)
};

\end{tikzpicture}

\caption{
\textbf{A.} Training time for 5{,}000 optimization steps as a function of the target dimensionality
(gene-space $G\in\{300,800,2000\}$ vs.\ program-space with fixed $K{=}20$); gene-space at $G{=}2000$
uses a reduced batch size (96 vs.\ 256).
\textbf{B.} Sampling time (seconds per 1{,}000 diffusion steps), showing the sharp increase in gene-space sampling
time with increasing $G$ and the near-constant program-space sampling time.
}
\label{fig:runtime_scaling_2panels}
\end{figure}

\subsubsection{Quantitative performance}\label{sec:quant_perf}
In this work, we address two practical limitations of diffusion-based morpho-transcriptomic inference: (i) sensitivity to normalization, and technical sequencing bias in the target space, and (ii) poor scalability when predicting high-dimensional gene vectors. We therefore evaluate (1) an analytically grounded normalization based on Pearson residuals, and (2) a program-based representation learned with consensus non-negative matrix factorization (cNMF), which constrains predictions to structured transcriptional factors and reduces the effective output dimension.

\paragraph{Impact of normalization.}
Table~1 shows the impact of normalization on prediction performance: for the same \textsc{STEM-2000} gene-space model, log-transformed targets yield higher Top-M PCC than Pearson-residual targets. This is consistent with prior work reporting higher predictive correlations under log-normalization; these higher correlations may partly reflect sequencing-depth- and cellularity-related signals retained by log-transformed representations \cite{ruyter_normalization_nodate}. We therefore adopt Pearson residuals as a controlled prediction target.

\paragraph{Program-space vs gene-space diffusion}
Under a fixed normalization (Pearson), program-space diffusion yields modest but consistent improvements over the STEM gene-space baseline across the reported Top-$M$ thresholds (Table~1).
Table~1 also reports a \textit{cNMF Oracle NNLS} in Pearson space, used as a reconstruction upper bound rather than as a predictive model. For slides from the held-out patient, true Pearson-normalized expression is projected onto the training-fold cNMF basis using NNLS and reconstructed in gene space. This estimates the best reconstruction achievable with the fixed basis: a low oracle indicates a representation bottleneck, whereas a large model--oracle gap indicates that predicting program activations from histology remains the main bottleneck.

\paragraph{Performance increases with panel size.}
Figure~\ref{fig:violin_top50_paper} shows that increasing the gene panel size leads to significantly better per-slide predictive quality (Top-50 PCC), with clear shifts in the distribution as we move from $G{=}300$ to $G{=}800$ and $G{=}2000$. Importantly, similar results are observed for both the gene-space (STEM) and the program-space (\textit{Ours}) diffusion methods. This trend motivates pushing toward larger panels to improve individual-level predictions. 

\paragraph{Computation scalability.}
We finally compare the computational cost of \textsc{STEM} in gene-space to program-space (predicting $K=20$ programs) while varying the gene panel size $G\in\{300,800,2000\}$.  In the gene-space, computation cost scales poorly with $G$, especially at inference: sampling throughput drops from 27.3\,it/s at $G{=}300$ to 7.4\,it/s at $G{=}800$ ($\sim$2\,min\,15\,s per 1{,}000 diffusion steps) and 1.7\,it/s at $G{=}2000$ ($\sim$36\,min per 1{,}000 steps). Training time also increases (5{,}000 steps: $\sim$25\,min at $G{=}300$, $\sim$1\,h\,31\,min at $G{=}800$, and $\sim$1\,h\,23\,min at $G{=}2000$, using batch size 96 instead of 256). In contrast, program-space diffusion keeps a fixed output dimensionality and is nearly constant across $G$: $\sim$8\,min per 5{,}000 training steps and 213--227\,it/s at sampling ($\approx$4.3--4.4\,s per 1{,}000 steps). The one-time cNMF extraction increases with $G$ (2/6/22 min for $G=300/800/2000$) but remains smaller than gene-space training time. Although the $G=2000$ training-time comparison is affected by the reduced batch size, the sampling results show that program-space diffusion scales favorably to larger gene panels while maintaining competitive predictive performance.

\section{Conclusion}
Normalization is a first-order determinant of reported accuracy: replacing STEM’s log normalization with Pearson residuals shifts Top-M PCC by changing the prediction target representation. Beyond preprocessing, larger gene panels improve per-slide predictive quality, but gene-space diffusion becomes prohibitively expensive at inference. By contrast, diffusion in a cNMF program space yields comparable or better Top-M performance under controlled normalization while making diffusion sampling nearly independent of the number of predicted genes. 
The current study remains limited to a single HER2+ cohort and does not establish cross-cohort generalization or biological utility beyond predictive agreement. Future work should evaluate broader cohorts, panel-specific K selection, and downstream tasks assessing biological utility beyond PCC-based metrics. 
Within this HER2+ cohort, our findings suggest that transcriptional program space provides a more structured and computationally scalable generative target than individual genes for morphology-to-transcriptomics prediction.

\subsubsection{Acknowledgments}

The research leading to these results has received funding from Agence Nationale de la Recherche as part of the ``France 2030'' program (reference ANR-23-IACL-0008, PRAIRIE-PSAI) and as part of the "Investissements d'avenir" program (reference ANR-19-P3IA-0001, PRAIRIE 3IA Institute; and  reference ANR-10-IAIHU-0006). The ARAMIS Lab is affiliated with DIM C-BRAINS, funded by the Conseil Régional d’Ile-de-France. This work was performed using HPC resources from GENCI–IDRIS (Grant 2025-AD011016416). R.D. received a Marie Skłodowska-Curie grant No 101154248 (project: SafeREG).

\subsubsection{Disclosure of Interests}
The authors have no competing interests in the paper 
%
%
%
\bibliographystyle{splncs04}
\bibliography{MICCAI2026}

\end{document}